\documentclass{article}
\usepackage{spconf,amsmath,graphicx}
\usepackage{amssymb,amsfonts}
\usepackage{cite}
\usepackage{booktabs}
\usepackage{url}
\usepackage{balance}
\usepackage{subfigure}
\def\x{{\mathbf x}}														
\def\q{{\mathbf q}}														
\def\z{{\mathbf z}}														
\def\C{{\mathbf C}}														
\def\W{{\mathbf W}}														
\def\ii{{\hat{\imath}}}												
\def\ij{{\hat{\jmath}}}												
\def\ik{{\hat{\kappa}}}												
\def\bH{\mathbb{H}}														
\def\bQ{\mathbb{Q}}														
\def\bR{\mathbb{R}}														
\def\cD{{\cal D}}															
\def\cL{{\cal L}}															
\def\cN{{\cal N}}															
\newcommand{\beps}{\boldsymbol{\epsilon}}					  
\newcommand{\bmu}{\boldsymbol{\mu}}					  
\newcommand{\bphi}{\boldsymbol{\phi}}				  
\newcommand{\bth}{\boldsymbol{\theta}}				
\newcommand{\bsigma}{\boldsymbol{\sigma}} 		
\newcommand{\bSigma}{\boldsymbol{\Sigma}} 		
\newcommand{\tT}{^{\text{T}}} 								
\newcommand{\tH}{^{\text{H}}} 								
\DeclareMathOperator{\diag}{diag} 									
\DeclareMathOperator{\E}{E} 									
\DeclareMathOperator{\Tr}{Tr}								
\newtheorem{definition}{Definition}
\title{A QUATERNION-VALUED VARIATIONAL AUTOENCODER}
\name{Eleonora Grassucci, Danilo Comminiello\thanks{Corresponding author's email: danilo.comminiello@uniroma1.it. This work has been supported by ``Progetti di Ricerca Grandi'' of Sapienza University of Rome under grant number RG11916B88E1942F.}, and~Aurelio~Uncini}
\address{Dept. Information Engineering, Electronics and Telecommunications (DIET)\\
Sapienza University of Rome, Via Eudossiana 18, 00184 Rome, Italy}
\begin{document}
%
\maketitle
%
\begin{abstract}
Deep probabilistic generative models have achieved incredible success in many fields of application. Among such models, variational autoencoders (VAEs) have proved their ability in modeling a generative process by learning a latent representation of the input. In this paper, we propose a novel VAE defined in the quaternion domain, which exploits the properties of quaternion algebra to improve performance while significantly reducing the number of parameters required by the network. The success of the proposed quaternion VAE with respect to traditional VAEs relies on the ability to leverage the internal relations between quaternion-valued input features and on the properties of second-order statistics which allow to define the latent variables in the augmented quaternion domain. In order to show the advantages due to such properties, we define a plain convolutional VAE in the quaternion domain and we evaluate its performance with respect to its real-valued counterpart on the CelebA face dataset.
\end{abstract}
%
%
%
%
%
%
\section{Introduction}
\label{sec:intro}
Variational autoencoders (VAEs) gained their success due to their ability of modeling a generative process formalized in the framework of probabilistic graphical models with latent variables \cite{KingmaARXIV2014, RezendeICML2014}. VAEs are characterized by a simple and fast sampling and easily accessible networks, which have proliferated their use in various applications. Recently, advanced VAEs have been developed, relying on statistical challenges \cite{RazaviICLR2019, VahadatICML2020} or on hierarchical architectures \cite{MaaloeNIPS2019, VahdatARXIV2020}.

Performing learning operations in the quaternion domain, rather than in the real-valued domain, has proved to provide higher efficiency when dealing with multidimensional data \cite{BulowTSP2001, MandicSPL2011, OrtolaniSIGPRO2017, ComminielloICASSP2019b}. Quaternions are widely used in neural networks since they allow to process groups of features together, thus coding latent inter-dependencies between features with a lower number of parameters with respect to real-valued neural networks \cite{GaudetIJCNN2018, ParcolletICLR2019, ParcolletAIR2019}. These advantages are due to the properties of quaternion algebras, including the Hamilton product that is used in quaternion convolutions. This has recently paved the way to the development of novel deep quaternion neural networks \cite{GaudetIJCNN2018, ParcolletAIR2019, VecchiTIT2020}, often tailored to specific applications, including theme identification in telephone conversation \cite{ParcolletTASL2020}, 3D sound event localization and detection \cite{ComminielloICASSP2019a, RicciardiMLSP2020}, heterogeneous image processing \cite{ParcolletICASSP2019a} and speech recognition \cite{ParcolletICASSP2019b}. Other properties of quaternion algebra that may be exploited in learning processes are related to the second-order statistics. In particular, the concept of properness for quaternion valued random signals and their applications has been widely investigated in the signal processing literature \cite{ViaTIT2010, TookSIGPRO2011, ViaTSP2011, OrtolaniMLSP2016, LeBihanSIGPRO2017}. 

In this paper, we investigate the concept of properness extended to the latent space in order to derive a quaternion-valued VAE (QVAE). Designing VAEs in the quaternion domain may bring several advantages. Neural networks for VAEs should model long-range correlations in data \cite{SadeghiARXIV2019, VahadatICML2020} and quaternion convolutional layers may provide additional information by leveraging internal latent relations between input features. Moreover, reducing the number of parameters may benefit the decoder and the marginal log-likelihood, which only depends on the generative network \cite{VahadatICML2020}.

The contribution of this paper is twofold: i) we define how to perform variational inference in the quaternion domain and ii) propose a plain convolutional QVAE. Moreover, to the best of our knowledge, for the first time augmented quaternion second-order statistics are exploited for the development of a deep learning model.

The paper is organized as follows. In Section~\ref{sec:qlearning}, we review the main learning operations in the quaternion domain. The proposed QVAE is derived in Section~\ref{sec:qvae} and then evaluated in Section~\ref{sec:results}. Finally, conclusions are drawn in Section~\ref{sec:conclusion}.
%
%
%
%
%
\section{Learning in the Quaternion Domain}
\label{sec:qlearning}
%
%
\subsection{Main Properties of Quaternion Algebra}
\label{subs:qalg}
The quaternion domain $\bH$ is defined by a four-dimensional associative normed division algebra over real numbers, belonging to the class of Clifford algebras. A quaternion is expressed by one real scalar component and three imaginary ones:

\begin{equation} 
	q = q_a + q_b\ii + q_c\ij + q_d\ik = q_a + \overline{q}.
	\label{eq:quaternion_definition}
\end{equation}

\noindent with $q_a, q_b, q_c, q_d \in \bR$. The imaginary units, $\ii=\left( 1,0,0 \right)$, $\ij=\left( 0,1,0 \right)$, $\ik=\left( 0,0,1 \right) $, are unit axis vectors and represent an orthonormal basis in ${\bR}^3$. The vector $\overline{q} = q_b\ii + q_c\ij + q_d\ik$ represents the imaginary part of the quaternion and it is also known as \textit{pure quaternion}. The imaginary units comply with the following fundamental properties:
\begin{gather}
	\ii^2 = \ij^2 = \ik^2 = -1,\\
	\ii \ij = \ii \times \ij = \ik, \quad \ij \ik = \ij \times \ik = \ii, \quad \ik \ii = \ik \times \ii = \ij,
	\label{eq:fundamprop}
\end{gather}

\noindent where ``$\times$'' denotes the vector product in ${{\mathbb{R}}^{3}}$. The above properties allow to adopt quaternion algebra $\bH$ to represent spatial rotations in $\bR^4$. Quaternion algebra $\bH$ is endowed with the operations of associative multiplication of elements in the algebra and scalar multiplication. The scalar product of two quaternions $p$ and $q$ is defined as $q \cdot p = q_a p_a + q_b p_b + q_c p_c + q_d p_d$. However, quaternions are not commutative under the operation of vector multiplication, thus $\ii \ij \ne \ij \ii$, and consequently $\ii \ij = -\ij \ii$, $\ij \ik = -\ik \ij$, $\ik \ii = -\ii \ik$. 

Due to the noncommutative property, the quaternion product, also known as Hamilton product, can be expressed as:
\begin{equation}
	\begin{split}
		qp &= \left(q_a + q_b\ii + q_c\ij + q_d\ik\right)\left(p_a + p_b\ii + p_c\ij + p_d\ik\right) \\
		&= \left(q_a p_a - q_b p_b - q_c p_c - q_d p_d\right) \\
		&+ \left(q_a p_b + q_b p_a + q_c p_d - q_d p_c\right)\ii \\
		&+ \left(q_a p_c - q_b p_d + q_c p_a + q_d p_b\right)\ij \\
		&+ \left(q_a p_d + q_b p_c - q_c p_b + q_d p_a\right)\ik. 
	\end{split}
	\label{eq:qprod}
\end{equation}

We can define the conjugate and the module of a quaternion, respectively, as $q^{*} = q_a - q_b\ii - q_c\ij - q_d\ik$ and $\left|q\right| = \sqrt{q_a^2 + q_b^2 + q_c^2 + q_d^2} = \left|q^*\right|$. A quaternion can be also written in polar form as $q = \left|q\right|\left(\cos\left(\theta\right) + \nu\sin\left(\theta\right)\right) = \left|q\right|e^{\nu\theta}$, where $\theta \in \bR$ is the argument of the quaternion, $\cos\left(\theta\right) = q_a/\left\|q\right\|$, $\sin\left(\theta\right) = \left\|\overline{q}\right\|/\left\|q\right\|$ and $\nu = \overline{q}/\left\|\overline{q}\right\|$ is a pure unit quaternion. The involution of a quaternion $q$ over a pure unit quaternion $\nu$ is $q^{\nu} = -\nu q \nu$, and it represents a rotation of $\pi$ in the imaginary plane orthogonal to $\left\{1, \nu\right\}$.

We will denote by boldface letters quaternions whose components are vectors (lowercase letters) or matrices (capital letters) of the same dimensions.
%
%
%
%
%
\subsection{Second-Order Statistics and Properness of Quaternion Random Vectors}
\label{subs:qsos}
Very often, it is necessary to study the characteristics of a quaternion signal by analyzing the second-order statistics. However, second-order information within a quaternion vector $\q$ cannot be estimated only from its correlation matrix $\C_{\q\q} = \E\left\{\q\q\tH\right\}$, but we need to define the complementary covariance matrices that augment the information within the covariance: $\C_{\q\q^\ii} = \E\left\{\q{\q^\ii}\tH\right\}$, $\C_{\q\q^\ij} = \E\left\{\q{\q^\ij}\tH\right\}$, $\C_{\q\q^\ik} = \E\left\{\q{\q^\ik}\tH\right\}$ (see also \cite{TookSIGPRO2011}, among others, for further details). Thus, we can introduce the augmented covariance matrix of an augmented quaternion vector $\tilde{\q} = \left[{\begin{array}{*{20}c}{\q\tT} & {{\q^{\ii}}\tT} & {{\q^{\ij}}\tT} & {{\q^{\ik}}\tT} \\ \end{array}} \right]\tT$, as:
\begin{equation}
	\tilde{\C}_{\q\q} = \E\left\{\tilde{\q}\tilde{\q}\tH\right\} 
	= \left[{\begin{array}{*{20}c} {{\C}_{\q\q}} & {{\C}_{\q\q^\ii}} & {{\C}_{\q\q^\ij}} & {{\C}_{\q\q^\ik}}\\
	{{\C}_{\q\q^\ii}\tH} & {{\C}_{\q^\ii \q^\ii}} & {{\C}_{\q^\ii \q^\ij}} & {{\C}_{\q^\ii \q^\ik}}\\
	{{\C}_{\q\q^\ij}\tH} & {{\C}_{\q^\ij \q^\ii}} & {{\C}_{\q^\ij \q^\ij}} & {{\C}_{\q^\ij \q^\ik}}\\
	{{\C}_{\q\q^\ik}\tH} & {{\C}_{\q^\ik \q^\ii}} & {{\C}_{\q^\ik \q^\ij}} & {{\C}_{\q^\ik \q^\ik}}\\
	\end{array}} \right]
	\label{eq:augcovariance}
\end{equation}

	%
It is now possible to introduce the quaternion-valued second-order circularity, or $\bQ$-properness.
\begin{definition}[\textbf{$\bQ$-Properness}]
	A quaternion random vector $\q$ is $\bQ$-proper iff the three complementary covariance matrices $\C_{\q\q^{\ii}}$, $\C_{\q\q^{\ij}}$ and $\C_{\q\q^{\ik}}$ vanish:
	\begin{equation}
		\E\left\{\q{\q^{\ii}}\tH\right\} = \mathbf{0} \quad \E\left\{\q{\q^{\ij}}\tH\right\} = \mathbf{0} \quad \E\left\{\q{\q^{\ik}}\tH\right\} = \mathbf{0}.
		\label{eq:qprop}
	\end{equation}
	\label{def:qproperness}
\end{definition}

\noindent The properness implies that $\q$ is not correlated with its vector involutions $\q^{\ii}$, $\q^{\ij}$, $\q^{\ik}$. The main properties of a $\bQ$-proper random variable are collected in Table \ref{table:qpropRV} \cite{TookSIGPRO2011}, where $\sigma^2$ denotes the variance of $\q$.
\begin{table}[t]
	\renewcommand{\arraystretch}{1.5}
	\caption{Properties of $\bQ$-proper random variables.}
	\label{table:qpropRV}
	\vspace{0.3cm}
	\begin{center}
		\begin{tabular}{ll}
			\hline
			$\E\left\{\q_\delta^2\right\} = \E\left\{\q_\epsilon^2\right\} = \sigma^2$ & $\forall \delta, \epsilon = \left\{a, b, c, d\right\}$\\
			$\E\left\{\q_\delta \q_\epsilon\right\} = \mathbf{0}$ & $\forall \delta, \epsilon = \left\{a, b, c, d\right\}$, $\delta \neq \epsilon$\\
			$\E\left\{\q \q\right\} = 
			-2 \sigma^2$ & $\forall \delta = \left\{a, b, c, d\right\}$\\
			$\E\left\{\left|\q\right|^2\right\} = 
			4 \sigma^2$ & $\forall \delta = \left\{a, b, c, d\right\}$\\
			\hline
		\end{tabular}
	\end{center}
\end{table}

A quaternion-valued random variable is Gaussian if all its components are jointly normal (see \cite{TookSIGPRO2011}, among others), i.e., $p\left(\tilde{\q}\right) = p\left(\q,\q^{\ii},\q^{\ij},\q^{ik}\right)$, thus the Gaussian probability density function for an augmented multivariate quaternion-valued random vector $\tilde{\q}$ is expressed as:
\begin{equation}
		p\left(\tilde{\q}\right) 
	= \frac{\exp\left\{-\frac{1}{2}\left(\tilde{\q} - \tilde{\bmu}\right)\tH\tilde{\C}_{\q\q}^{-1}\left(\tilde{\q} - \tilde{\bmu}\right)\right\}}{\left(\pi/2\right)^{2N} \det\left(\tilde{\C}_{\q\q}\right)^{1/2}},
	\label{eq:pdfqaug}
\end{equation}

\noindent where $\tilde{\bmu}$ is the augmented quaternion-valued mean vector for $\tilde{\q}$. For a $\bQ$-proper random vector, using the properties of Table \ref{table:qpropRV} and replacing \eqref{eq:qprop} in \eqref{eq:augcovariance}, the augmented covariance matrix $\tilde{\C}_{\q\q}$ becomes equal to $4 \sigma^2 \mathbf{I}$ (see also \cite{TookSIGPRO2011}), and the mean value refers directly to the quaternion vector, i.e., $\bmu$, thus we obtain a simplified expression of the Gaussian distribution:
\begin{equation}
	p\left(\tilde{\q}\right) = \frac{1}{\left(2\pi\sigma^2\right)^{2N}}\exp\left\{-\frac{1}{2\sigma^2}\left(\q - \bmu\right)\tH\left(\q - \bmu\right)\right\}
	\label{eq:pdfqaugprop}
\end{equation}

\noindent where the argument of the exponential is a real function of only $\left|\q - \bmu\right|^2 = \left(\q - \bmu\right)\tH\left(\q - \bmu\right)$.
%
%
%
%
%
%
\section{Variational Autoencoder in the Quaternion Domain}
\label{sec:qvae}
Here, we introduce the novel quaternion variational autoencoder (QVAE) as a generative method based on the probabilistic relation between the quaternion-valued input space and the quaternion-valued latent space. The QVAE aims at controlling the distribution of the quaternion-valued latent vector, which has the characteristic of being a $\bQ$-proper quaternion-valued random vector.
%
%
%
%
%
\subsection{Variational Inference in the Quaternion Domain}
\label{subs:qvinf}
Let us consider a quaternion-valued latent vector $\z \in \bH$, characterized by a prior probability distribution $p_{\bth}\left(\z\right)$, and a quaternion input $\x \in \bH$, whose conditional probability density function with respect to $\z$ is expressed as $p_{\bth}\left(\x|\z\right)$. Similarly to the real-valued VAE \cite{KingmaARXIV2014}, the QVAE introduces an approximation $\hat{p}_{\bphi}\left(\z|\x\right)$ of the true posterior distribution, thus it is possible to express the marginal likelihood as:
\begin{align}
		&\log\left(p_{\bth}\left(\x\right)\right) = \lambda \cD_{\text{KL}}\left(\hat{p}_{\bphi}\left(\z|\x\right) || p_{\bth}\left(\z|\x\right)\right) + \cL\left(\bth, \bphi; \x\right)\nonumber\\
		&= - \lambda \cD_{\text{KL}}\left(\hat{p}_{\bphi}\left(\z|\x\right) || p_{\bth}\left(\z\right)\right) + \E_{\hat{p}}\left\{\log\left(p_{\bth}\left(\x|\z\right)\right)\right\}
	\label{eq:costfunction}
\end{align}

\noindent where $\cD_{\text{KL}}\left(\cdot\right)$ denotes the Kullback-Leibler (KL) divergence and $\cL\left(\bth, \bphi; \x\right)$ is the variational lower bound with respect to $\bth$ and $\bphi$ \cite{BleiICML2012, KingmaARXIV2014}. The parameter $\lambda$ is used to better scale KL values with respect to $\cL\left(\bth, \bphi; \x\right)$.

We assume that both the recognition model $\hat{p}_{\bphi}\left(\z|\x\right)$ and the generative model $p_{\bth}\left(\z\right)$ are based on $\bQ$-proper Gaussian distributions, i.e., they follow the rule in \eqref{eq:pdfqaugprop}. In particular, the distribution of $\hat{p}_{\bphi}\left(\z|\x\right)$ can be characterized by a quaternion-valued mean vector $\bmu_{\z}$ and by an augmented covariance matrix $\tilde{\C}_{\z\z}$. This matrix is a block-diagonal matrix, whose sub-matrices on the main diagonal, extending the properties of the $\bQ$-proper random variables in Table~\ref{table:qpropRV}, are equal and show the same variance, thus $\tilde{\C}_{\z\z} = \diag\left\{\bSigma_{\z}, \bSigma_{\z}, \bSigma_{\z}, \bSigma_{\z}\right\}$, where $\bSigma_{\z} = \diag\left\{\bsigma_{\z}^2\right\}$ and $\bsigma_{\z}^2$ is the quaternion-valued vector containing the variance values for each quaternion component. The quaternion-valued mean and variance vectors, $\bmu_{\z}$ and $\bsigma_{\z}^2$ respectively, are computed by using each a quaternion-valued single-layer neural network. 

On the other hand, the prior $p_{\bth}\left(\z\right)$ is assumed to be a centered isotropic $\bQ$-proper Gaussian distribution, i.e., $p_{\bth}\left(\z\right) \backsim \cN\left(\z; \mathbf{0}, \mathbf{I}\right)$. These considerations allow to approximate the expectation in \eqref{eq:costfunction} as $\E_{\hat{p}}\left\{\log\left(p_{\bth}\left(\x|\z\right)\right)\right\} \approxeq \frac{1}{L} \sum_{l=1}^{L}$ ${\log \left(p_{\bth}\left(\x|\z_l\right)\right)}$, where $\z_l$, with $l=1,\ldots,L$, denotes the first $L$ samples drawn from $\hat{p}_{\bphi}\left(\z|\x\right)$.

Once created the $\bQ$-proper Gaussian prior distribution, we perform a reparametrization trick similarly to the real-valued VAE \cite{KingmaARXIV2014}, 
thus having $\z = \bmu_{\z} + \bsigma_{z}\odot\beps$, where $\odot$ denotes an element-wise product and $\beps \in \bH \backsim \cN\left(\mathbf{0},\mathbf{I}\right)$.
%

Considering that both $\hat{p}_{\bphi}\left(\z|\x\right)$ and $p_{\bth}\left(\z\right)$ are $\bQ$-proper distributions and that $\tilde{\C}_{\beps\beps} = 4\mathbf{I}$, $\bmu_{\beps} = \mathbf{0}$ being $p_{\bth}\left(\z\right) \backsim \cN\left(\z; \mathbf{0}, \mathbf{I}\right)$, the quaternion-valued KL loss in \eqref{eq:costfunction} can be expressed as \cite{ViaTIT2010, DoerschARXIV2016}:
\begin{align}
		&\cD_{\text{KL}}\left(\hat{p}_{\bphi}\left(\z|\x\right) || p_{\bth}\left(\z\right)\right) = \frac{1}{2}\left(\Tr\left\{\tilde{\C}_{\beps\beps}^{-\frac{1}{2}}\tilde{\C}_{\z\z}\tilde{\C}_{\beps\beps}^{-\frac{1}{2}}\right\}\right. \nonumber\\
		&+ \left. \left(\bmu_{\beps} -\bmu_{\z}\right)\tH\tilde{\C}_{\beps\beps}^{-1}\left(\bmu_{\beps} -\bmu_{\z}\right) - N + \log\left(\frac{\det\left(\tilde{\C}_{\beps\beps}\right)}{\det\left(\tilde{\C}_{\z\z}\right)}\right)\right)\nonumber \\
		&= \frac{1}{2}\left(\Tr\left\{\bSigma_{\z}\right\} + \bmu_{\z}\tH\bmu_{\z} - N\right) - 2\sum_{i=1}^N{\log\left(\sigma_{\z i}^2\right)}.
		\label{eq:kldqvae}
\end{align}

\noindent It is worth noting that the minimization of the KL divergence \eqref{eq:kldqvae} provides a measure of $\bQ$-improperness \cite{ViaTIT2010}.
%
%
%
%
%
\subsection{Network Architecture}
\label{subs:qnet}
For the scope of the paper, here we use a rather simple architecture in order to prove the benefits of the variational inference in the quaternion domain. Thus, we consider an encoder network composed of 
quaternion convolutional layers. 

The quaternion convolution is one of the main operations of deep neural networks in the quaternion domain \cite{ParcolletAIR2019, GaudetIJCNN2018}. Considering a generic quaternion input vector, $\q$, defined similarly to \eqref{eq:quaternion_definition}, and a generic quaternion filter matrix defined as $\W = \W_{a} + \W_{b}\ii + \W_{c}\ij + \W_{d}\ik$, the quaternion convolution can be expressed as the following Hamilton product:
\begin{equation}
	\begin{split}
		\W \otimes \q &= \left(\W_{a}\q_{a} - \W_{b}\q_{b} - \W_{c}\q_{c} - \W_{d}\q_{d}\right)\\
		&+ \left(\W_{a}\q_{b} + \W_{b}\q_{a} + \W_{c}\q_{d} - \W_{d}\q_{c}\right)\ii \\
		&+ \left(\W_{a}\q_{c} - \W_{b}\q_{d} + \W_{c}\q_{a} + \W_{d}\q_{b}\right)\ij \\
		&+ \left(\W_{a}\q_{d} + \W_{b}\q_{c} - \W_{c}\q_{b} + \W_{d}\q_{a}\right)\ik \\
	\end{split}
	\label{eq:hamprod}
\end{equation}

\noindent The Hamilton product of \eqref{eq:hamprod} allows quaternion neural networks to capture internal latent relations within the features of a quaternion. Each quaternion convolutional layer is followed by 
a split quaternion Leaky-ReLU activation function \cite{ParcolletAIR2019}. Quaternion batch normalization is not taken into account as it may be considered as a source of randomness that could cause instability \cite{MaaloeNIPS2019, VahdatARXIV2020}. Then, as also said in the previous subsection, a quaternion-valued fully-connected output layer is added to the encoder to achieve the quaternion-valued augmented covariance matrix that is used to sample the quaternion latent variable $\z$ and compute the KL divergence loss.

Similarly to the encoder, for the decoder network we use quaternion transposed convolutional layers, each one followed by 
a quaternion Leaky-ReLU activation function.
%
%
%
%
%
\section{Experimental Results}
\label{sec:results}
In this section, we evaluate the proposed method on the CelebFaces Attributes Dataset (CelebA) \cite{LiuICCV2015}, a large-scale face attribute dataset with $202,599$ images that we crop and scale to $64 \times 64$ pixels, as in \cite{HouWACV2017}. We investigate the behavior of the proposed QVAE in comparison with a plain VAE in both reconstruction and generation tasks. We report generated samples from both the models for a visual evaluation, as well as results in terms of structural similarity index measure (SSIM), mean-square error (MSE) and Fr\'echet Inception Distance (FID), for a more objective appraisal.
\begin{table}[t]
	\caption{Averaged results from objective metrics on reconstruction (SSIM, MSE) and generation (FID) tasks. Scores can be read as follows: the higher the better for SSIM, the lower the better for MSE and FID.}
	\label{table:res_metrics}
	\vspace{0.3cm}
	\centering
	\begin{tabular}{@{}lllll@{}}
		\toprule
		& SSIM            & MSE             & FID		& \# parameters          \\ \midrule
		VAE  & 0.8492          & 0.0047          &195.7		& 3,762,539          \\
		QVAE & \textbf{0.8941} & \textbf{0.0031} &\textbf{175.7} 	& \textbf{1,404,996} \\ \bottomrule
	\end{tabular}
\end{table}

We consider an encoder with $5$ convolutional blocks for the VAE and quaternion-convolutional layer for the proposed QVAE, both with dimensions $(32, 64, 128,$ $256, 512)$, and a transposed convolutional decoder $(512, 256,$ $128, 64, 32)$ (equivalently, quaternion transposed convolutions in the proposed method). Both the networks consider Adam optimizer with an initial learning rate of $0.0005$ decreased by a factor of $0.5$, similarly as \cite{HouWACV2017}. The dimensionality of the latent space 
is set to $100$ in the quaternion domain\footnote{The implementation of the QVAE is available online at \url{https://github.com/eleGAN23/QVAE}.}. We consider a loss composed of binary cross entropy (BCE) for reconstruction and a KL divergence weighted by $\lambda = 0.00001$.

Due to the quaternion operations, the QVAE network has an incredibly lower number of parameters with respect to the real-valued VAE (about less than half of the parameters, as shown in Table \ref{table:res_metrics}), thus gaining considerable memory advantages. However, despite the lighter architecture in terms of parameters, the proposed QVAE clearly outperforms the real-valued method in terms of objective metrics for the reconstruction task. As shown in Table \ref{table:res_metrics}, the QVAE scores a significantly better value both for SSIM and for MSE.

\begin{figure}
	\centering
	\includegraphics[keepaspectratio,width=\columnwidth]{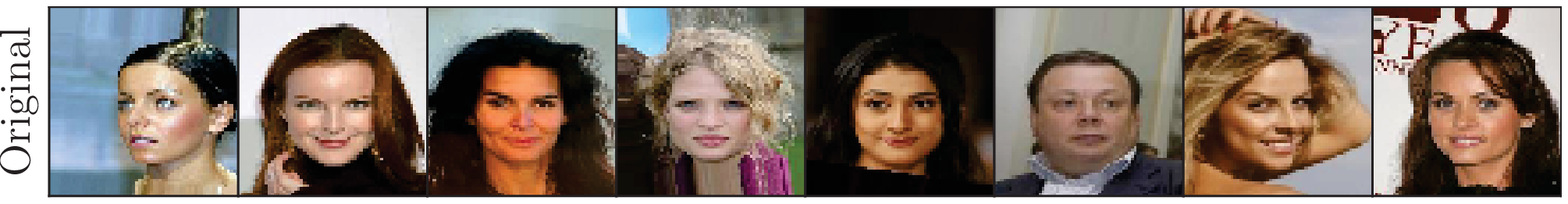}\\
	\includegraphics[keepaspectratio,width=\columnwidth]{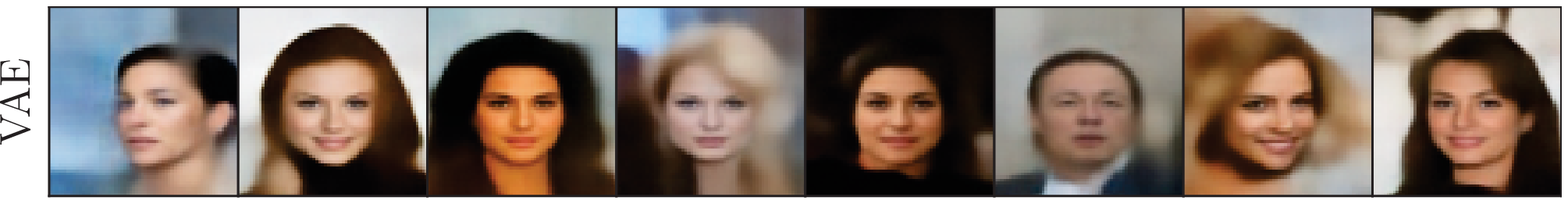}\\
	\includegraphics[keepaspectratio,width=\columnwidth]{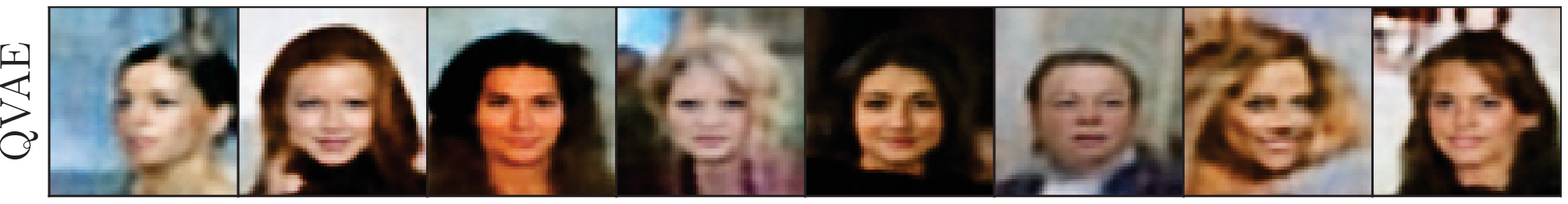}
	\caption{Original test set and reconstructed samples sets from plain VAE and proposed QVAE.}
	\label{fig:recons_img}
\end{figure}

Figure \ref{fig:recons_img} reports the reconstructed images compared with the original ones. On one hand, VAE generates samples focusing just on the face and leaving hair, neck and background very blurred, almost indistinguishable. On the other hand, the proposed QVAE is able to generate samples considering both face and background. Indeed, even hair and background are reconstructed more similarly to the ground truth samples, thus producing more realistic images overall.

\begin{figure}
	\centering
	\includegraphics[keepaspectratio,width=\columnwidth]{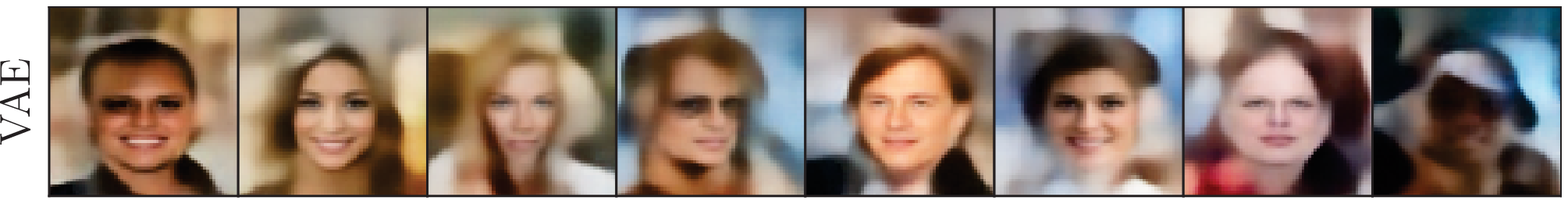}\\
	\includegraphics[keepaspectratio,width=\columnwidth]{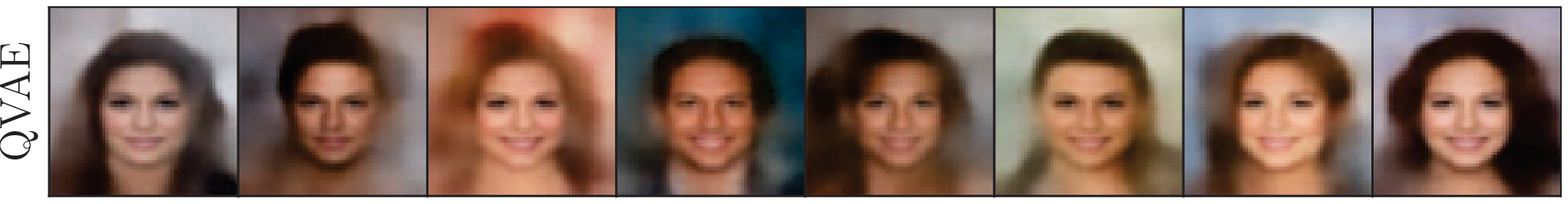}
	\caption{Generated fake image samples from the plain VAE and the proposed QVAE.}
	\label{fig:gen_img}
\end{figure}

Concerning the generation, which is usually the most challenging task, we report both the generated sample sets and the results in terms of  objective metric. Figure \ref{fig:gen_img} shows the generated images from the plain VAE and the ones sampled from the proposed QVAE network. While the first set seems to have a more various background, the generated faces are less detailed and sometimes confused with the environment. On the contrary, the set sampled from the QVAE shows less heterogeneous background, but significantly more accurate face contours and, overall, a stabler generation. The superior generation ability of the proposed QVAE is underlined also by the results in Table \ref{table:res_metrics} in terms of the FID score. The FID computes the distance of the statistics of generated samples with real ones, thus lower FID values correspond to more real generated samples.

Both reconstruction and generation results prove the effectiveness of defining a latent space and operating in the quaternion domain, while having a huge reduction of the overall number of network parameters.

%
%
%
%
%
\section{Conclusion}
\label{sec:conclusion}
In this paper, we proposed a novel approach for VAEs by defining the generative model in the quaternion domain. In particular, the proposed proper QVAE is able to learn latent representations in the quaternion domain by leveraging the augmented second-order statistics of the quaternion-valued input. Moreover, the QVAE involves the quaternion convolutional layers in both the encoder and the decoder networks, which lead to an impressive reduction of the overall number of network parameters. We considered a plain QVAE to clearly show the benefits of the new quaternion-based approach. Results have shown the effectiveness of the proposed approach in both reconstruction and generation tasks with respect to the real-valued counterpart. Future works will extend the proposed QVAE approach to the improper case and to more complex and advanced deep generative models.

%
%
\balance
\bibliographystyle{IEEEbib}
\ninept
\bibliography{QVAE_arXiv}
\end{document}